\documentclass[conference]{IEEEtran}
\usepackage{mathptmx} %
\usepackage{amsmath,amssymb,amsfonts}
\usepackage{algorithmic}
\usepackage[ruled,vlined]{algorithm2e}
\usepackage{graphicx}
\usepackage{textcomp}
\usepackage{xcolor}
\usepackage{fancyhdr}
\usepackage{textcomp}
\usepackage{tabularx}
\usepackage{adjustbox}
\usepackage{threeparttable}
\usepackage{float}
    \newcolumntype{L}{>{\raggedright\arraybackslash}X}
\usepackage[caption=false]{subfig}
\usepackage{bm}

\def\BibTeX{{\rm B\kern-.05em{\sc i\kern-.025em b}\kern-.08em
    T\kern-.1667em\lower.7ex\hbox{E}\kern-.125emX}}

\renewcommand{\baselinestretch}{0.93}

\newcommand{\ignore}[1]{}

\newcommand{\arxiv}[1]{} %

\newcommand{\nameofwork}{YOSO}

\newif\ifremark
\long\def\remark#1{
\ifremark%
        \begingroup%
        \dimen0=\columnwidth
        \advance\dimen0 by -1in%
        \setbox0=\hbox{\parbox[b]{\dimen0}{\protect\em #1}}
        \dimen1=\ht0\advance\dimen1 by 2pt%
        \dimen2=\dp0\advance\dimen2 by 2pt%
        \vskip 0.25pt%
        \hbox to \columnwidth{%
                \vrule height\dimen1 width 3pt depth\dimen2%
                \hss\copy0\hss%
                \vrule height\dimen1 width 3pt depth\dimen2%
        }%
        \endgroup%
\fi}
\remarktrue

\usepackage{comment}

\usepackage{catoptions}
\makeatletter

\def\Autoref#1{%
  \begingroup
  \edef\reserved@a{\cpttrimspaces{#1}}%
  \ifcsndefTF{r@#1}{%
    \xaftercsname{\expandafter\testreftype\@fourthoffive}
      {r@\reserved@a}.\\{#1}%
  }{%
    \ref{#1}%
  }%
  \endgroup
}
\def\testreftype#1.#2\\#3{%
  \ifcsndefTF{#1autorefname}{%
    \def\reserved@a##1##2\@nil{%
      \uppercase{\def\ref@name{##1}}%
      \csn@edef{#1autorefname}{\ref@name##2}%
      \autoref{#3}%
    }%
    \reserved@a#1\@nil
  }{%
    \autoref{#3}%
  }%
}
\makeatother

\usepackage{fancyhdr}
\usepackage[normalem]{ulem}
\usepackage[sort,nocompress]{cite}
\usepackage[final]{microtype}
\usepackage[keeplastbox]{flushend}

\usepackage[bookmarks=true,breaklinks=true,letterpaper=true,colorlinks,linkcolor=black,citecolor=blue,urlcolor=black]{hyperref}
\usepackage{cleveref}

\newcommand\T{\rule{0pt}{2.275ex}}       %
\newcommand\B{\rule[-1.05ex]{0pt}{0pt}} %

\title{You Only Spike Once: Improving Energy-Efficient Neuromorphic Inference to ANN-Level Accuracy}

\author{\IEEEauthorblockN{Srivatsa P\IEEEauthorrefmark{1}\IEEEauthorrefmark{3}, Kyle Timothy Ng Chu\IEEEauthorrefmark{1}\IEEEauthorrefmark{3}, Burin Amornpaisannon\IEEEauthorrefmark{1}\IEEEauthorrefmark{3}, Yaswanth Tavva\IEEEauthorrefmark{1},\\ Venkata Pavan Kumar Miriyala\IEEEauthorrefmark{1},
Jibin Wu\IEEEauthorrefmark{2}, Malu Zhang\IEEEauthorrefmark{2}, Haizhou Li\IEEEauthorrefmark{2} and Trevor E. Carlson\IEEEauthorrefmark{1}}
\IEEEauthorblockA{\IEEEauthorrefmark{1}School of Computer Science \IEEEauthorrefmark{2}Department of Engineering\\
National University of Singapore}%
\thanks{\emph{This work has been submitted to the IEEE for possible publication. Copyright may be transferred without notice, after which this version may no longer be accessible.}}%
\thanks{\emph{\IEEEauthorrefmark{3}S. P, K. T. N. Chu and B. Amornpaisannon contributed equally to this work.}}%
}

\IEEEoverridecommandlockouts
\begin{document}
\maketitle

\pagestyle{plain}

\begin{abstract}
In the past decade, advances in Artificial Neural Networks (ANNs) have allowed them to perform extremely well for a wide range of tasks. In fact, they have reached human parity
when performing image recognition, for example.
Unfortunately, the accuracy of these ANNs comes at the expense of a large number of cache and/or memory accesses and compute operations. Spiking Neural Networks (SNNs), a type of neuromorphic, or brain-inspired network, have recently gained significant interest as power-efficient alternatives to ANNs, because they are sparse, accessing very few weights, and typically only use addition operations instead of the more power-intensive multiply-and-accumulate (MAC) operations. The vast majority of neuromorphic hardware designs support rate-encoded SNNs, where the information is encoded in spike rates. Rate-encoded SNNs could be seen as inefficient as an encoding scheme because it involves the transmission of a large number of spikes. A more efficient encoding scheme, Time-To-First-Spike (TTFS) encoding, encodes information in the relative time of arrival of spikes. While TTFS-encoded SNNs are more efficient than rate-encoded SNNs, they have, up to now, performed poorly in terms of accuracy compared to previous methods.

Hence, in this work, we aim to overcome the limitations of TTFS-encoded neuromorphic systems.
To accomplish this, we propose: (1) a novel optimization algorithm for TTFS-encoded SNNs converted from ANNs and
(2) a novel hardware accelerator for TTFS-encoded SNNs, with a scalable and low-power design.

Overall, our work in TTFS encoding and training improves the accuracy of SNNs to achieve state-of-the-art results on MNIST MLPs, while reducing power consumption by 1.46$\bm{\times}$ over the state-of-the-art neuromorphic hardware.

\end{abstract}

\section{Introduction}
In recent years, Artificial Neural Networks (ANNs) have demonstrated excellent results in a wide range of real-world computational problems such as object detection, speech recognition and image classification. ANNs have been improving in accuracy, and in 2015, crossed an important threshold, beating human accuracy~\cite{DBLP:journals/corr/HeZR015} at the ImageNet 2012 Visual Recognition Challenge~\cite{imagenet_2012}. However, the effectiveness of ANNs comes at the cost of high power consumption. In short, the performance of these networks relies on an extremely large number of model parameters, requiring a huge number of computational resources. This tends to make large ANNs unsuitable for low-powered applications such as Internet-of-Things (IoT) and mobile devices. To address this issue, there has been an increased focus on developing energy efficient networks, including EfficientNet~\cite{efficientnet_2019}, MobileNet~\cite{2020_Howard_Mobilenetv3} and SqueezeNet~\cite{squeezenet_2016}, to meet the efficiency needs of these low-power systems.

While these more efficient networks are promising for deployment to low powered devices, the use of spiking neural networks (SNNs) allows for even greater power savings. In SNNs, information is represented by binary events called spikes, similar to the way information is communicated in the human brain. This technique of mimicking brain functionality, called neuromorphic computing, makes use of only addition operations, instead of multiply-and-accumulate operations in standard ANNs, which has the capability to significantly reduce the computational power and complexity needed. Furthermore, SNNs can also take advantage of the sparsity of their neuron activations to significantly reduce the movement of data. As a result, SNNs have garnered significant interest over the last few years as a candidate for low-powered accelerators~\cite{Merolla668}.

While there are several encoding methods for SNNs~\cite{ponulak_introduction_nodate}, the two most prominent ones are rate-based~\cite{2014_gerstner_Leaky_IF, snn_single_neurons_populations_plasticity_2002} and temporal encoding~\cite{temporal_encoding_1995, temporal_encoding_2004}.  Because temporally encoded networks have not been able to match the state-of-the-art accuracy of rate-based coded networks~\cite{encoding_comparison_2018}, rate-based encoding has been the most common encoding scheme used in hardware SNN accelerators~\cite{loihi_2018, minitaur_2014}.
In rate-based encoding, information is represented by the average number of spikes over a period of time, and the inference results become more accurate as additional spikes are generated.
Because each spike triggers memory accesses (to load parameter information) which need to be fetched from on- or off-chip memory, the power consumption in rate-encoded networks can be relatively high. Hence, some works have turned to temporal encoding instead to further take advantage of the sparsity of the networks~\cite{mostafa_fast_2017}. One type of temporal encoding is known as time-to-first-spike (TTFS) encoding. Under this encoding, information is represented by the relative time of arrival of the spikes with respect to the first spike, not the average number of spikes over a time period.
Previously, TTFS encoding had traded off power efficiency for accuracy, unable to match the results of rate-encoded systems. In this work, we propose a training method to leverage the power efficiency TTFS encoded SNNs with minimal loss to accuracy.

Among existing rate-based accelerators, IBM's TrueNorth is the most well known. With the ability to hold up to 1 million neurons and 256 million synapses, and can consume 65mW when running typical computer vision applications~\cite{akopyan_truenorth:_2015}, TrueNorth still remains the state-of-the-start in terms of power efficiency. However, while TrueNorth is extremely power efficient at implementing rate-based networks, it is unable to take full advantage of the properties of temporally encoded networks. One of the main reasons behind TrueNorth's low power usage is that the number of memory accesses on each core does not scale with the number of spikes it receives and therefore remains constant. While having a constant number of memory accesses per tick works well for rate-based networks which produce large numbers of spikes, it prevents TrueNorth from maximizing efficiency of temporally encoded networks with sparse spikes, as the memory access rate does not decrease with this sparsity. Thus, this work introduces the You Only Spike Once (\nameofwork{}) accelerator, a novel architecture specifically designed to leverage the sparsity in the spiking behavior of temporally coded networks.

One of the largest barriers preventing the widespread adoption of SNNs is the fact that SNNs are notoriously difficult to train from scratch. While significant progress has been made in recent years~\cite{ihmehmelli_2019}, their results are still far behind what has been achieved by state-of-the-art ANNs today for classification problems. To circumvent this problem, we instead chose to convert pre-trained ANN models into SNNs before mapping them onto the proposed hardware. Previous work~\cite{rueckauer_conversion_2018} has introduced a technique that can convert ANNs to SNNs with minimal loss in accuracy for small networks. For larger and deeper networks, this approach does not work as well. Hence we have developed a novel training method that improves the accuracy of the converted SNNs through this technique.

The main goal of this paper is to run highly accurate networks on low power devices, with minimal loss to accuracy. Therefore, this work focuses on (1) optimizations to improve the accuracy of TTFS-encoded SNNs, and (2) optimizations to the hardware running the network. Hence, the contributions of this work are:

\begin{itemize}
    \item An end-to-end neuromorphic technique that demonstrates state of the art performance and accuracy for TTFS-encoded SNNs
    \item A new training algorithm that reduces the approximation error which accumulates as a result of converting ANNs to SNNs. In doing so, our TTFS-encoded SNNs reach near ANN-accuracy (within 0.2\%), allowing TTFS-encoded SNNs to be considered for traditional ANN tasks, at much higher efficiency. 
    \item An implementation of a novel hardware accelerator for TTFS-encoded SNNs that is configurable and scalable. Our work significantly improves power efficiency with minimal reduction in the accuracy of network performance.
\end{itemize}

\section{Background} \label{background}

Spiking neural networks (SNNs) have garnered significant interest over the last few years, primarily as a candidate for energy-efficient inference on low-powered devices. In SNNs, information is represented by discrete binary events called spikes, similar to the way the brain represents information. This is unlike a standard artificial neural network (ANN) where information is represented as continuous values~\cite{taherkhani_2020_snn_learning_review}. The use of SNNs reduces the computational power needed by only requiring cheaper addition operations compared to the more power-intensive
multiply-and-accumulate (MAC) operations used in ANNs. Furthermore,
SNNs typically have an extremely low activation count,
in comparison to their non-spiking counterparts. Activations are low %
because every neuron is only activated by a strictly positive input, a subset of all possible inputs, above a pre-defined threshold. This translates to just a small subset of all neurons firing for any given inference. A small subset of neurons firing translates into a low memory access count and, hence, a low cost when performing inference tasks. %

SNNs are fundamentally different from ANNs. ANNs traditionally involve a synchronous tensor multiplication for each layer while SNNs involve an asynchronous propagation of information. The way information can be propagated through an SNN can vary. The two most prominent methods are rate-based~\cite{snn_single_neurons_populations_plasticity_2002, 2014_gerstner_Leaky_IF} and temporal encoding~\cite{temporal_encoding_1995, temporal_encoding_2004}. In rate-based encoding, information is represented by the mean firing rate of the neurons. Although there exist different definitions of firing rate, it often denotes either spikes averaged over repetitions of an experiment or the average number of spikes over a period of time. This work refers to the latter when referring to rate-encoded networks. Rate-encoded networks become more accurate over time as more spikes are generated. From a power consumption point of view, each spike will require a weight look-up. Because rate-based encoding has many spikes, having a weight look-up for each spike limits the minimum number of memory accesses and the corresponding amount of energy saved.

An alternative form of encoding uses temporal encoding which is based on spike timing~\cite{mostafa_fast_2017}. Common temporal encoding schemes include Time to First Spike (TTFS), where information is represented by the relative time of arrival of the spikes with respect to the first spike, and phase-of-firing, where information is encoded using the time at which neurons fire within a periodic cycle~\cite{cattani2015phaseoffiring}. When information is encoded in the TTFS scheme, neurons in an SNN spike at most once during each inference pass and see many fewer spikes compared to their rate-based counterparts. By definition, the rate encoding scheme relies on the generation of multiple spikes over a fixed period of time, while the TTFS encoding scheme relies on the time taken for a single neuron to spike. Therefore, the TTFS encoding scheme allows for fewer spikes compared to a rate-based encoding scheme. Assuming a spike corresponds to a memory access, the TTFS encoding scheme allows for a low number of memory accesses. In addition, an inference pass of a TTFS network can end once the output layer produces its first output spike instead of waiting for the rest of the inputs to arrive. As a result, a minimal number of computations are performed for any particular inference, making temporal encoding a highly suitable candidate for encoding energy-efficient SNNs.

\subsection{Formalizing TTFS-SNNs}
SNNs have been proposed to model the biological neural network of brains that use spikes to represent and communicate information across neurons~\cite{snn_single_neurons_populations_plasticity_2002}. As the fundamental information processing units in the biological neural networks, the spiking neurons are composed of dendrite, soma, and axon. Dendrites receive weighted inputs from the preceding neurons, which are further integrated into the membrane potential of the soma. An output spike is generated from the soma once the membrane potential crosses the firing threshold. The output spike is then transmitted to the subsequent neurons through the axonal connections. A number of spiking neuron models have been proposed to describe the internal dynamics and diversified characteristics of biological neurons. One such model is the Hodgkin-Huxley~\cite{1952_hodgkin_hodgkin_huxley_model} model which describes sodium and potential channels in a biological neuron using differential equations. Such a model is needlessly complicated. Instead, the reduced Spike Response Model~(SRM)~\cite{2001_gerstner_spike_response_model} is typically used. Integrate-and-fire(IF), Leaky integrate-and-fire~(LIF)~\cite{2014_gerstner_Leaky_IF}, and Exponential integrate-and-fire~\cite{2003_fourcaud_exponential_IF} models are variants of SRM. In these models, neuron behaviour is modelled by a membrane potential that integrates incoming current. When the accumulated membrane potential of a neuron crosses a threshold, the neuron spikes. The integration of input currents is described by a kernel function and occurs when a neuron is connected to another through a synapse. The order in which multiple  such neurons are connected produces different spiking neural network topologies.

In this work, to properly encode information into spike timings, we use a non-leaky IF neuron model~\cite{rueckauer_conversion_2018}. The membrane potential dynamics of this model can be described by the following equation:
\begin{equation}
\frac{dV^i_{mem}(t)}{dt} = \sum\limits_j w_{ij}\sum\limits_n\kappa_{ij}(t-t_j^n) + b_{i}t
\label{eq1:model_neuron}
\end{equation}
where $V^i_{mem}$ is the membrane potential of neuron $i$, and $w_{ij}$ is the weight of the synaptic connection from the pre-synaptic neuron $j$ to the post-synaptic neuron $i$. $t_j^n$ is the timing of the $n$th spike from the pre-synaptic neuron $j$. Since in TTFS-encoding we are concerned with the time associated with only the first spike ($n=1$), hereon we ignore subsequent spikes and refer to the time of the first spike of neuron $j$ using $t_j$. $\kappa_{ij}$ is the kernel that describes the induced post-synaptic potential (PSP) by the incoming spikes.
To preserve the spike timings within the sub-threshold membrane potential, the PSP kernel is chosen to use the piece-wise linear function defined as follows:

\begin{equation}
\label{eq:psp}
\kappa(t-t_j) = [t-t_j]\Theta(t-t_j)
\end{equation}

where $\Theta$ is the heaviside step function defined as:
\begin{equation}
\label{eq:heaviside_function}
    \Theta(x) = \begin{cases}
   1 & \text{if \quad $x\geq 0$}  \\
   0       & \text{otherwise} 
  \end{cases}
\end{equation}

The heaviside step function $\Theta$ can be ignored by only considering input spikes that arrive before $t_i$ for each neuron $i$. Because it is these input spikes that influence the output spike of each neuron $i$, we consider the pre-synaptic neurons that produce these input spikes as a set of causal neurons. The set of causal neurons $\Gamma_i^<$, can hence be defined as $\Gamma_i^< := \{j | t_j < t_i\}$. Hence, the Eq. (\ref{eq1:model_neuron}) can be simplified into
\begin{equation}
\label{eq:model_neuron_2}
\frac{dV^i_{mem}(t)}{dt} = \sum\limits_{j \in \Gamma_i^<}  w_{ij}[t-t_j] + b_{i}t
\end{equation}

In order to determine the time to first spike for each neuron $i$, we set the membrane potential to its threshold value $\theta$:

\begin{equation}
\label{eq:model_neuron_3}
\theta = \sum\limits_{j \in \Gamma_i^<}  w_{ij}[t_i-t_j] + b_{i}t_i
\end{equation}

By re-ordering the terms of the Eq. (\ref{eq:model_neuron_3}), the time to first spike for the neuron $i$ can be expressed as follows
\begin{equation}
\label{eq:model_neuron_7}
t_i = \frac{1}{\mu_i}\left(\theta + \sum\limits_{j \in \Gamma_i^<} w_{ij}t_j\right)
\end{equation}
where 
\begin{equation}
\label{eq:model_neuron_8}
\mu_i := \sum\limits_{j \in \Gamma_i^<}  w_{ij}  + b_{i}. 
\end{equation}

As the instantaneous firing rate $r_i$ of the neuron $i$ is the inverse of $t_i$, the proposed ANN-to-SNN conversion method~\cite{rueckauer_conversion_2018} equates activation $a_i$ in an ANN to the instantaneous rate $r_i$ of the corresponding neuron $i$ in the converted SNN, assuming the use of ReLU activation functions in the ANN. The state of SNNs have changed over the years, and have shown significant progress.

\section{Related Work} \label{related-work}

There are three key approaches to achieve power efficient neural network inference covered by this work. They include spiking neural networks, hardware accelerators and neural network optimizations.

\subsection{Spiking Neural Networks}

SNNs can be constructed by either training from scratch or converting from a pre-trained ANN. Although many spike-based training algorithms~\cite{shrestha2018slayer, wu2019direct, 2019_mostafa_surrogate_gradient_learning} have shown promising results on the MNIST~\cite{rueckauer_conversion_2018} dataset, these algorithms have not been tested rigorously on larger network architectures and more challenging datasets. As an alternative to SNN-based training, pre-trained ANNs can be converted into SNNs. This method has been shown to be highly successful for rate-encoded SNNs on the CIFAR-10 and ImageNet~\cite{2017_rueckauer_conversion_imagenet} datasets. These previous works have focused on rate-encoded SNNs, where a large number of synaptic operations are required. %
As a result, rate-based encoding greatly limits the power efficiency %
of SNN models when deployed onto the neuromorphic hardware.

Compared to rate-encoded SNNs, temporally encoded SNNs are able to run with fewer operations and memory accesses~\cite{mostafa_fast_2017} which are highly desirable attributes for low-powered devices (details in Section \ref{background}). Although significant power savings can be achieved by using TTFS-encoded SNNs instead of rate-encoded SNNs, TTFS-encoded SNNs that are constructed by either training from scratch~\cite{ihmehmelli_2019, mostafa_fast_2017, DBLP:journals/corr/Mostafa16, zhang2020spiketimingdependent} or converting from the pre-trained SNN~\cite{rueckauer_conversion_2018} did not perform as well as their ANN counterparts in terms of the classification accuracy. As demonstrated in a recent study~\cite{rueckauer_conversion_2018}, converting from ANNs to TTFS-encoded SNNs, unfortunately, leads to accumulated approximation errors, which results in significantly lower accuracy in the SNNs as compared to the equivalent ANNs, particularly in larger network architectures. Our work tackles this problem by proposing a novel training approach to refine the network weights after conversion, which improve the performance of the converted TTFS-encoded SNNs. 

\subsection{Hardware Accelerators}

In the past few years, there have been several architectures that have been proposed for neuromorphic hardware. The most prominent among them is IBM's TrueNorth neuromorphic chip~\cite{akopyan_truenorth:_2015}, which has an extremely low power density of just 20 milliwatts per square centimeter and has shown results equivalent to state of the art on several benchmarks. However, TrueNorth does not take advantage of sparse activations to reduce the number of memory accesses. Regardless of the number of spikes, cores on the TrueNorth chip will always perform a read for each neuron in their core SRAM. Such an implementation can be attributed to the usage of non-standard networks for inference, making no assumptions about the networks that are run. As a result, TrueNorth needs to handle connections on a per-neuron basis, requiring the hardware to keep track of the connections of every single neuron. There is a high overhead for such an implementation, where half of the data used during read operations (256 out of 410 bits) on TrueNorth are to check for connectivity.

Unlike TrueNorth, our work implements standard networks with an exploitable access pattern. Hence, we are able to make reasonable assumptions about neuron connectivity and can express these connections as a layer-wise access pattern instead of storing them individually for each neuron. As a result, we are able to perform significantly fewer reads per time step (See Table~\ref{table:truenorthcomp}).

\begin{table}[t]
\caption{Read and write comparison of our work with TrueNorth~\cite{akopyan_truenorth:_2015}. Numbers in bytes. r=Reads, w=Writes.}
\label{table:truenorthcomp}
\resizebox{\columnwidth}{!}{%
\begin{tabular}{l|r|r|r|r} 
\hline
\hline
                \T\B & r (spike) & w (spike) & r (timestep) & w (timestep) \\
\hline
 TrueNorth \T      & -                 & 1/8                & 13,312               & 13,312                   \\
 Our Work (MLP) \B  & 768               & 512                & 1024                  & 512                   \\
\hline
\end{tabular}
}%
\end{table}

In the space of Time-To-First-Spike (TTFS) based hardware accelerators, the viability of  Time-To-First-Spike based systems %
have been demonstrated by examining the sparsity of TTFS-encoded networks~\cite{mostafa_fast_2017}. However, the work did not explore the potential for a low powered accelerator, as the authors have not %
leveraged the sparsity of activations in a TTFS-encoded SNN in a significant way. Furthermore, the accuracy reported was far below the state-of-the-art.

Other notable SNN architectures include Intel's Loihi~\cite{loihi_2018} and Minitaur~\cite{minitaur_2014}, which feature online training, and a processor for Deep Neural Networks with Binary/Ternary Weights in 28nm CMOS~\cite{BNNprocessor}.

\subsection{Quantization}

Quantization is one technique that is often applied to SNNs, with layers having 4- or 8-bit precision~\cite{2019_bouvier_neuromorphic_hardware}.
In this work, we examine the effects of 8-bit quantization on efficiency and accuracy.

\subsection{Summary}

Power efficient inference can be achieved through the use of extremely sparse SNNs. While most works use rate-encoded SNNs instead of TTFS-encoded SNNs, trading accuracy for power efficiency, we show that is possible to achieve greater power efficiency for comparable accuracy, through the use of TTFS-encoded SNNs. %

In this work, we demonstrate a novel hardware accelerator specifically designed for TTFS-encoded SNNs.
Along with the energy-efficient accelerator, we propose a method to convert ANNs to SNNs which allows our TTFS-encoded SNNs to leverage network compression techniques for more power savings.
In this work, we propose the combination of enhanced training and efficient hardware to demonstrate the potential of TTFS-based platforms.

\section{Training Competitive TTFS-SNNs}
TTFS-encoded SNNs have shown better power efficiency and inference speed as compared to their rate-based variants. The converted TTFS-encoded SNNs, however, suffer from quantization errors that accumulate across layers. This significantly deteriorates the classification accuracy, particularly in deeper SNNs, as compared to their equivalent ANNs. Another source of error arises when an input spike, coming from the synaptic connection with a large weight, drives a neuron's internal membrane potential across the firing threshold, before subsequent inhibitory input spikes that targeting the same post-synaptic neuron arrives. This problem can be explained by the different operating mechanism of the spiking neuron and artificial neuron, wherein the input information is distributed and integrated over time by the SNN rather than at the same time instant as happened in the ANNs. Raising the threshold value of the post-synaptic neurons may alleviate this problem. However, it is not a good option in practice since it adversely increases the latency for decision making.

To address these problems, we propose a training method to systematically convert pre-trained ANNs to the TTFS-encoded SNNs. First, we apply a data-driven weight normalization strategy such that the neuron activation is not dominated by a few input spikes with large weights while also ensure timely decision making. Finally, to mitigate conversion errors, we propose a layerwise training methodology. As a whole, the proposed training framework effectively closes the accuracy gap between the pre-trained ANNs and the converted SNNs.

\subsection{Firing Threshold Determination}
Determining the right combination of neuronal firing threshold, weight and bias values is crucial to striking a balance between the classification accuracy and latency. Apart from the learnable parameters (weights and biases) that can be directly taken from the pre-trained ANNs, the firing threshold requires extra effort to be determined. An inappropriate threshold value will cause the converted SNN to perform significantly poorer compared to the equivalent ANN. One common approach to this problem would be to set the threshold to 1 and adjust the weights such that the activations are normalized.  

\subsection{Weight Normalization}

In order to prevent the converted SNNs from underestimating output activation of the corresponding ANNs, this work applies weight normalization. One way to normalize weights is to consider all possible combinations of positive activations that could occur at a particular ANN layer and scale the weights by that maximum quantity. The benefit of such an approach is that it only depends on the weights and biases of the network. However, in reality the maximum activation that determined in this way might be far from the actual activation values for majority of neurons. This leads to weights and biases that are much smaller than they need to be, increasing the time taken for a neuron to get activated. Because the time taken for a neuron to first spike increases, a longer duration will be required to achieve high classification accuracy. This problem will be exacerbated in deeper networks if weights are normalized in this way for all layers.

Instead of this conservative approach, we estimate the maximal activation values of an ANN by making use of the training data~\cite{2015_diehl_snn_threshold_weight_balancing}. Images from the training set are converted into input spike trains before being propagated through a $L$-layered ANN to produce $n$ sets of $L$ activation vectors $\{\bold{a^1}...\bold{a^L}\}$ where each set contains the activation vectors obtained from propagating each of the $n$ input spike trains. The scale factor for each layer, $s_{current}$, is set to the magnitude of the maximum activation observed in $\{\bold{a^1}...\bold{a^L}\}$. Weight and bias vectors are then scaled by the scale factor. Note that because this algorithm uses data from the training set, a strong performance guarantee cannot be extended to the test set. As long as the training and test sets have a similar data distribution, which is typically the case, the activation vectors observed using the training set would be similar to that observed in the test set. However, there might be instances where the activations observed in test set are more extreme than those observed in training set, potentially leading to a reduction in accuracy. Nevertheless, the benefit of this method over the former method is that it provides a much better trade off between latency and accuracy. This is because the time taken to spike is shorter to achieve a similar accuracy. 

\subsection{Training Network}

Errors arising from converting ANNs to SNNs can be further reduced through (1) retraining an ANN with constraints or (2) refining the learnable parameters on the converted
temporally-encoded SNN. While retraining a standard ANN with constraints might be feasible for small tasks such as the MNIST dataset, it might be extremely challenging to do so with larger networks on larger tasks such as ImageNet.

Training algorithms typically applied to SNNs can be broadly classified into two categories - membrane potential driven and spike driven. Membrane potential driven training algorithms treat membrane potentials as differentiable signals and use surrogate derivatives~\cite{2019_mostafa_surrogate_gradient_learning}, or use Back-Propagation Through Time (BPTT) ~\cite{shrestha2018slayer, wu2019direct, 2017_wu_bptt}. While competitive results are shown~\cite{wu2019direct}, this class of algorithms can be extremely demanding in terms of memory and compute as the entire forward pass of the network needs to be stored in order to compute derivatives.

The second category is spike driven learning algorithms which rely on spike timings to train a network. These algorithms~\cite{2000_bohte_error_backpropagation, 2017_shrestha_spike_train_learning, 2020_hong_spike_training} usually assume that membrane potentials are linear around the time at which the neuron spikes to avoid dealing with non-differentiability. There are some works~\cite{DBLP:journals/corr/Mostafa16} that do not require this assumption by using IF neurons.

Instead of choosing between training SNNs from scratch and converting ANNs-to-SNNs, we propose to both convert an ANN to an SNN and train to minimize approximation errors. This allows for significant reduction in the training time of accurate SNNs. We propose coupling each layer in an ANN and the corresponding layer in the converted SNN, and minimzing a layer wise cost function. Unlike traditional SNN training algorithms which utilize a loss computed at the final layer, the algorithm we are proposing is aimed at minimizing the divergence between ANN activations $a_{Li}$ and SNN activations $s_{Li}$ for every neuron with index $i$ in a layer $L$.

From the ANN-SNN
conversion, the analog activation of a neuron in the ANN is equivalent to the
instantaneous firing rate of TTFS-encoded SNN. The instantaneous firing rate is given by the inverse of the time taken for a neuron to first spike. It is possible to model the approximation between the activation of a single neuron $i$ in a particular layer $l$ in an ANN and the corresponding neuron in an SNN: $a^l_i = \frac{1}{t^l_i} + \epsilon$ where the introduction of $\epsilon$ allows for activation between SNN and ANN to deviate by a reasonable margin of error. A potential loss function is the L2-norm, given by $L = \frac{1}{2}*(a^l_i - r^l_i)^2$ where $r_{li}$ is the instantaneous firing rate of neuron $i$ in layer $l$ given by $r^l_{i} = \frac{1}{t^l_{i}}$.

The loss function is minimized by updating synaptic weights as described in Algorithm \ref{alg:finetuning-algorithm}. For each layer, $L$, the divergence between the ANN activation vector and SNN instantaneous rate vector is computed and minimized. This has the effect of delaying or advancing spike times in the network. In Section~\ref{sec:results}, we demonstrate how this improved training method works to increase inference accuracy.   

The loss function is minimized by updating synaptic weights as described in Algorithm \ref{alg:finetuning-algorithm}. $\beta$ is the fraction of neurons to keep. By removing the least salient $1-\beta$ neuron weights in $get\_parameters\_from\_ann$, it is possible to remove connections that could potentially lead to long-latency spikes. For each iteration in all iterations $K$, the forward pass of ANN produces a set of activation vectors, $\{\bold{a^1}...\bold{a^L}\}$, and SNN produces a set of first time spikes vectors, $\{\bold{t^1}...\bold{t^L}\}$. For each layer, $L$, the divergence between the ANN activation vector and SNN instantaneous rate vector is computed and minimized. This has the effect of delaying or advancing spike times in the network.

\begin{algorithm}[t!] 
\textbf{Input:} $\{I^1...I^n\}$: Set of $n$ input spikes vectors generated from randomly sampling $n$ images from training set \\
\textbf{Input:} $\beta$: Fraction of neurons to keep in each layer\\
\textbf{Input:} $\eta$: Learning rate\\
\textbf{Input:} $\epsilon$: Margin of error \\
\textbf{Input:} $K$: Number of iterations \\
\textbf{Output:} Finetuned weights vector $\mathbf{w}$ \\
$\mathbf{w_i}$ $\leftarrow$ get\_parameters\_from\_ann($\beta$)\;
$\mathbf{w_n}$ $\leftarrow$ normalize\_weights($\mathbf{w_i}$, $\{I^1...I^n\}$)\;

k = 0

\While{k $<$ K and error $>$ $\epsilon$} { 
    \For{$I^r$ in $\{I^1...I^n\}$}{
        // Get activation vectors for each layer in an $L$-layered ann \\
        $\{\mathbf{a^1}...\mathbf{a^L}\}$ $\leftarrow$ ann\_forward\_pass($I^r$)\;
        // Get vectors of spike times for each layer in an $L$-layered snn \\
        $\{\mathbf{t^1}...\mathbf{t^L}\}$ $\leftarrow$ snn\_forward\_pass($I^r$)\;
        // Get vectors of instantaneous spike rates for each layer in an $L$-layered snn 
        $\{\mathbf{r^1}...\mathbf{r^L}\}$ $\leftarrow$ get\_spike\_rates($\{\mathbf{t^1}...\mathbf{t^L}\}$)\;
        \For{q=1 to L}{
            layer\_type $\leftarrow$ get\_layer\_from\_index(q)\;
            \uIf{layer\_type in \{batch norm, dropout\}}{
                skip
            }
            \Else{
                $error$ = $L_2(\mathbf{a^q}, r^q)$\;
            }
            // update weight \\
            $\mathbf{w_n}$ -= $ \eta*\frac{\partial L}{\partial w}*error $\;
        }
    }
    k+=1
}
\caption{\emph{train\_network}: SNN training}
\label{alg:finetuning-algorithm}
\end{algorithm}

\section{Architecture Description}
\subsection{Abstract Hardware Model}

This section describes an abstract hardware model which we used to translate the mathematical SNN models described in earlier sections into a model that is easier to translate into actual physical hardware.

The abstract model consists of a computational block that loads from and writes to three storage blocks and represents a single layer of neurons $i$. The hardware unit receives two forms of inputs - (1) incoming spikes from neurons $j$ in the previous layer and (2) End-of-Timestep~(EoT) packets used to signal the unit to move onto the next timestep $t+1$.

The Weight block stores the weights of the synaptic connections $w_{ij}$ while the Accumulated Weights block keeps track of the gradient of the neuron potential $\frac{dV^i_{mem}(t)}{dt}$ in equation \ref{eq1:model_neuron}. The Neuron Potentials block stores the neuron potential $V^i{mem}(t)$ as well as information on whether the neuron has already produced a spike.

The Computational Block contains a set of registers that are used to either keep track of the current state of layer or are used to generate various access patterns of different network types.

\subsubsection{Processing input spikes}

The core implements a spike processing algorithm which uses four registers to implement an access pattern. One register is used to keep track of the address to be accessed while another will be used to keep track of the number of accesses made. The remaining two registers are populated on program time and store the number of accesses $P$ that need to be made and the address increment after each access $M$. When a spike arrives at the core from neuron $j$, the address register will be set based on the index of $j$. The core will then generate $P$ memory accesses, incriminating the address by $M$ after each access. Using this algorithm, our core is able to implement fully connected networks efficiently, as unnecessary computations and memory accesses will be skipped over by the access pattern.

\subsubsection{Processing End-of-Timestep~(EoT) Signals}
In SNNs, time is used as an additional mechanism to store information.
In YOSO, EoT signals are used to indicate that a timestep is complete and the accelerator can move on to the next one.
Unlike the processing of input spikes, no additional information needs to
be decoded from the EoT signals. The Computational Block supports two ways of handling EoT signals - the standard Integrate-and-Fire method
and the softmax method
which is normally used in the final layer of the networks. \\

In the standard Integrate-and-Fire method, spikes are generated as long as the neuron potential crosses the threshold and the neuron has not spiked before. However, in the softmax method, only the neuron with the largest neuron potential produces a spike. In both methods, an EoT signal is sent to the next layer after the Computational Block has finished updating the neuron potentials and generating spikes.

\subsection{Architecture Description}
\begin{figure}[t]
\centering
\subfloat[A YOSO processing element]{%
  \includegraphics[trim={1cm 2.5cm 0cm 2.75cm},clip,width=0.85\columnwidth]{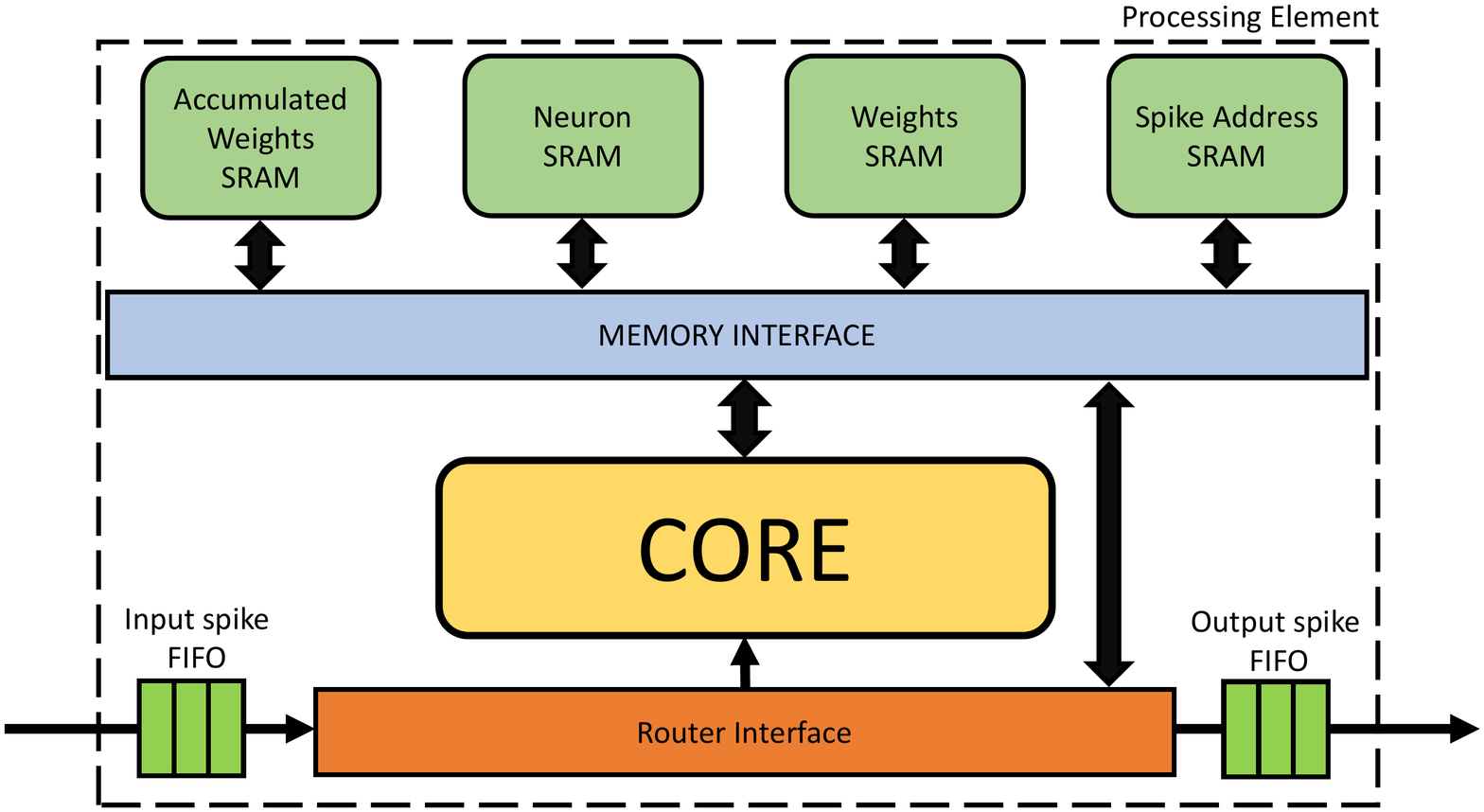}%
  \label{fig:top_level_processing_element}%
}

\subfloat[A detailed diagram of the core]{
\includegraphics[trim={0.7cm 2cm 1cm 2.5cm},clip,width=0.78\columnwidth]{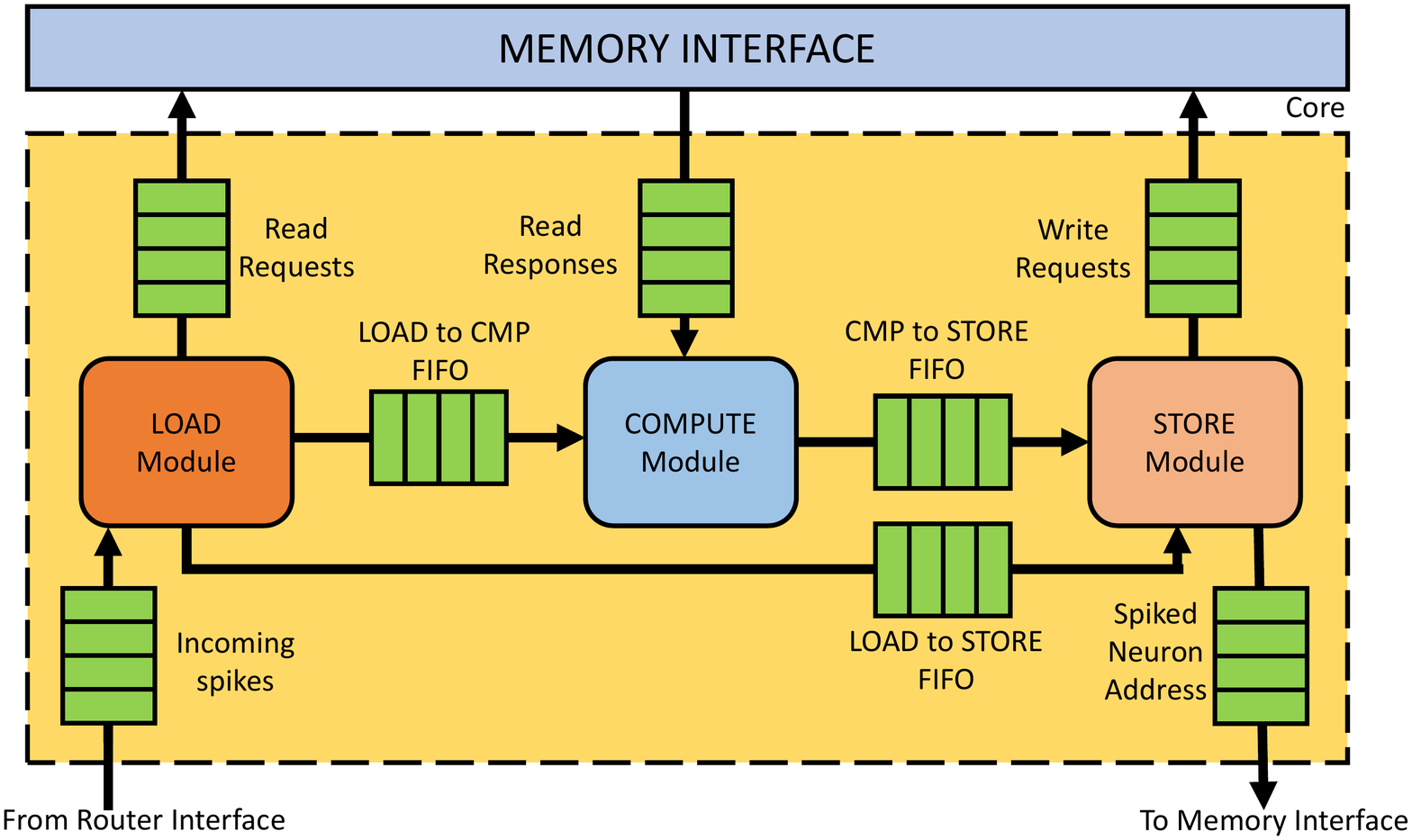}%
\label{fig:core}%
}
\caption{A processing element (a) and its core (b), the main components of the \nameofwork{} neuromorphic processor.}
\end{figure}

In this section, we discuss the detailed design of the YOSO accelerator. The accelerator consists of multiple Processing Elements (PEs) that are connected together through a Network-on-Chip (NoC) with each PE supporting up to 256 neurons. In this work we build on the OpenSMART NoC architecture~\cite{kwon2017opensmart} to implement a lightweight NoC that utilizes x-y routing to send spike packets from one PE to another. Figure \ref{fig:top_level_processing_element} shows the layout of the different components of a PE. Under normal operations, the core will update the accumulated weights and the neuron potentials by accessing and updating the appropriate data from the SRAM banks attached to the memory interface. 

\subsubsection{Router Interface}
The router interface is responsible for sending input spikes to the appropriate components depending on the current state of the PE. When the PE is in programming mode, incoming spikes may be sent to the memory interface to set the initial values of the SRAM blocks while all spikes will be directed to the core under normal operation.

The router interface also supports two methods of generating output packets that are sent to the router. First, the router may take the output spike from the Spike Address Storage -- when a spike is generated by the neuron core, a 8-bit neuron address is sent to the Spike Address Storage SRAM which sends the 32-bit spike to the router interface. The router interface then appends the 8-bit coordinates of the core \textit{output destination} and sends the 40-bit packet to the router. The second method of output packet generation is by forwarding received spikes. Spike forwarding allows a single layer to be mapped across multiple cores without the need for the sender to keep track of the coordinates of all the cores in the layer. If forwarding is active, the router interface sends spikes received from the router to the core while also using it to create a 40-bit packet by appending the 8-bit coordinates of the core \textit{forwarding destination} which is then sent back to the router. 

\subsubsection{Memory Interface}

The memory interface consists of four individual SRAM interfaces -- one for each of the SRAM blocks in the PE. As YOSO uses single-port SRAMs, only a single read/write request can be processed at one time. The SRAM interfaces alternate between servicing requests from the Write Request FIFO and the Read Request FIFO to ensure that all requests are processed in a reasonable amount of time.
SRAM blocks whose write queues are connected to the router interface are populated on program time while SRAM blocks with write queues connected to the core will be updated during run time.

\textbf{Handling RAW dependencies.}
To handle RAW dependencies, read requests must contain an additional bit which indicates if the read is done with the intention to alter the current value. Additionally, SRAM interfaces with RAW protection contain a special 256-bit RW Protection Register - 1 bit for each entry in the SRAM. When a read request with the intention to write occurs, the bit in the RW Protection Register indexed by the read address will be set to 1 and will only be set back to 0 after a write request to that same address is processed. Any subsequent reads (regardless of whether they intend to alter the current value) will be stalled until the matching write request is processed. Reads that do not have the intention to write will not cause the RW Protection Register bit to be set. Since the Weight SRAM is only written to during program time, its SRAM interface does not contain this mechanism as RAW dependencies will not occur during runtime.

\subsubsection{Core}

The core is the key computational element of the PE. The design of the core was inspired by traditional deep learning accelerators like VTA~\cite{moreau2018vta} and adopts a decoupled access-execute model~\cite{smith1982decoupled} to memory access hide latency. Each core consists of 3 modules that communicate with each other through FIFOs. This allows for a better utilization of the cores' resources as the other modules can continue execution if one of them encounters a stall.

\textbf{Load Module.}
The load module is a finite state machine with two states: an idle state and an active state. In the idle state, the load module waits for an input spike to arrive from the router interface. When an input spike arrives at the core, the information encoded in the spike is decoded and used to populate the initial values of the internal registers.
If a register is not set by the decoded spike values, they are set by the values of reference registers which are written to during the programming stage. After the register values have been set, the load module sends the spike type along with the number of addresses to be generated to the compute module and transits to the active state.

In the active state, the load module begins to generate %
read requests 
which are sent to different SRAM blocks depending on the spike type. %
The generated addresses are sent to both the memory interface and the store module. After all the necessary addresses are generated, the load module transits to the idle state and waits for the next input spike.

\textbf{Compute Module.}
Like the load module, the compute module is a finite state machine with an idle state and an active state. In the idle state, the compute module waits for the load module to send the spike type and the number of addresses to be generated. The compute module initializes its registers with this data and transits to the active state.

In the active state, the compute module performs addition operations on the data retrieved from the memory interface in response to the requests generated by the load module. Saturating adders are used to handle overflow and underflows. The results are then sent to the store module along with the spike type. After all the data is processed, the compute module transits back to the idle state.

\textbf{Store Module.}
The store module is responsible for checking for spikes and storing updated values back to the appropriate SRAM block. The store module takes the values from the compute module and stores them in the address obtained from the load module. Since there is no re-ordering of read requests, we can ensure that the values from the compute module and the store module correspond to one another. If the spike is an EoT signal, the store module will first check if the neuron potential crosses the threshold before storing the value. The store module supports spike generation for both TTFS and softmax layers. In order to support softmax, reserved spike types
are used to mark the first and last neuron in the layer. The store module then picks the neuron highest neuron potential within the range of neurons marked out by the two neurons and sends out the spike associated with the neuron, regardless of whether it has crossed the threshold. 

\section{Mapping}
This section, we explain how SNNs are mapped to the \nameofwork{} accelerator. In order to map a $m \times n$ fully connected layer, a minimum of $C = MAX(\frac{n}{N},\frac{m \times n}{W})$ PEs are needed where $N$ is the maximum number of neurons that can be mapped to a single core and $W$ is the maximum number of weights that the core can contain. The PEs are placed within a $\sqrt{C}$ by $\sqrt{C}$ grid. Each PE will be allocated to a layer to process all the spikes received by the layer throughout the inference stage. To maximise throughput and reduce latency, layers that contain more neurons and those that process more spikes will be allocated more PEs.

\textbf{Parameter Mapping.}
 Mapping a FCN to a \nameofwork{} core is a straightforward process. Biases are stored in the neuron SRAM as the initial value of the neuron potentials.

\section{Experimental Methodology}

In this section, we outline the details of the experimental setup and algorithms used in evaluating  our works.

\subsection{Input \& Output representations}

Benchmarking SNNs require input data to be encoded as spike trains. For visual datasets, possible techniques include using: (1) Event-based sensors - creating a dataset using event-based cameras to generate spike trains (2) Stochastic methods - conversion of image intensity of images from conventional datasets into Poisson/Bernoulli spike trains (3) Intensity to Latency (ItL) encoding - generating a spike train containing a spike per pixel in the image, where the spike's latency is inversely proportional to the intensity of a pixel in an image. The use of stochastic methods could potentially be useful in improving generalizability of networks at the cost of added training stage complexity. Hence, this work uses the simpler ItL encoding scheme for data from conventional image classification datasets used for ANNs. Although not difficult, the extension of this work to integrate input data from event-based sensors is a potential avenue for future work.

The output layer of our SNN implementation is one-hot encoded. The categorical classes of each dataset are encoded such that the index of the neuron that spikes corresponds to one of the output classes. 

\subsection{Networks}

The MNIST Handwritten Digits dataset~\cite{lecun-mnisthandwrittendigit-2010} contains grey-scale images of 10 handwritten digits of size 28 x 28, with a total training set of 60,000 examples, and a test set of 10,000 examples. We built a fully-connected network with three layers~(300-300-10) with hidden layers containing 300 neurons in the hidden layer.

\subsection{Hardware Simulation}

The YOSO accelerator was synthesized using Synopsys Design Compiler version P-2019.03-SP5 targeting a 22nm technology node with a $6\times7$ PE configuration. Gate-level simulation was performed using Synopsys VCS-MX K-2015.09-SP2-9 and power analysis was performed with Synopsys PrimePower version P-2019.03-SP5. The simulations were run at 120KHz and accounts for both programming and inference time. The system configuration used is showin in Table~\ref{tab:system_parameters}, and the per-PE SRAM configuration is shown in Table~\ref{tab:sram_config}.

\begin{table}[t]
    \caption{Hardware configuration parameters}
    \subfloat[SRAM Configuration (Per-PE)]{%
        \begin{tabular}{l|r}
        \hline
        \hline
        Name & \T\B Size  \\
        \hline
        Accumulated Weights \T & 1 kB  \\
        Neurons & 1 kB \\
        Weights & 40 kB \\
        Spike Address \B & 2 kB \\
        \hline
        \end{tabular}
        \label{tab:sram_config}
    }
    \hfill
    \subfloat[System Parameters]{%
        \begin{tabular}{l|r}
        \hline
        \hline
        Name & \T\B Size  \\
        \hline
        Tech node \T & 22nm \\
        Clock frequency & 120~kHz \\
        Cores & 42 \\
        NoC Dimensions & 6x7 \\
        Routing \B & X-Y \\
        \hline
        \end{tabular}
        \label{tab:system_parameters}
    }
\end{table}

\section{Results and Analysis}
\label{sec:results}

In this section, we first describe the accuracy, power and energy efficiency results, and finally analyze the choice of output layer used.

\subsection{Performance}

\begin{table}[ht]
\caption{Performance on MNIST Dataset (TTFS-encoding)} %
\label{table:mnist_results} %
\centering
\begin{threeparttable}
\begin{adjustbox}{width=0.48\textwidth}
\begin{tabular}{l c c c} 
\hline\hline %
Network \T\B & Coding & ANN acc(\%) & SNN acc(\%) \\ [0.5ex] %
\hline 
TrueNorth-a~\cite{esser2015backpropagation} \T & Rate & - & 92.70 \\
Rueckauer et al~\cite{2017_rueckauer_conversion_imagenet} & Rate & 98.56 & 98.50 \\
TrueNorth-b~\cite{esser2015backpropagation} & Rate & - & 99.42 \\
Mostafa~\cite{DBLP:journals/corr/Mostafa16} & Temporal & - & 97.55 \\
Comsa et al~\cite{ihmehmelli_2019} & Temporal & - & 97.96\\
Rueckauer et al~\cite{rueckauer_conversion_2018} & Temporal & 98.56 & 98.30  \\
\bf{YOSO (Our Work)} & Temporal & 98.56 & \bf{98.44} \\
\bf{YOSO (Our Work) + Quant.}\B & Temporal & 98.56 & \bf{98.40} \\
\hline %
\end{tabular}
\end{adjustbox}
\end{threeparttable}
\end{table}

\begin{table}[t]
        \caption{Comparison of this proposal with general neuromorphic accelerators on the MNIST dataset sorted by accuracy. Our work simultaneously demonstrates both low power and high accuracy. Enc. is encoding, Acc. is Top-1 accuracy in percent, fps is frames per second, Tech is in nm, power in mW.}
    \label{tab:comparison_to_scalable_hardware}
    \resizebox{\linewidth}{!}{%
    \begin{threeparttable}
    \begin{tabular}{l l c r r r r}
    \hline
    \hline
        \multicolumn{1}{l}{Accelerator\T\B} & Enc. & Acc. & \multicolumn{1}{c}{fps} & \multicolumn{1}{c}{Tech} & \multicolumn{1}{c}{Power} & \multicolumn{1}{c}{uJ/frame} \\ \hline
        SNNwt~\cite{du2015neuromorphic}\T\B & Rate & 91.82 & -& 65 & - & 214.700 \\ \hline
        TrueNorth-a~\cite{esser2015backpropagation}\T\B & Rate & 92.70 & 1000 & 28 & 0.268 & 0.268 \\ \hline
        Spinnaker~\cite{khan2008spinnaker}\T\B & Rate & 95.01 & 77 & 130 & 300.000 & 3896.000 \\ \hline
        Tianji~\cite{ji2018bridge}\T\B & Rate & 96.59 & - & 120 & 120.000 & - \\ \hline
        Shenjing~\cite{wang2019shenjing}\T\B & Rate & 96.11 & 40 & 28 & 1.260 & 38.000 \\ \hline
        \bf{YOSO~(this work)}\T\B & Temp. & 98.40 & 26 & 22 & \bf{(0.859*)~0.734} & \bf{(32.985*)~28.185} \\ \hline 
        TrueNorth-b~\cite{esser2015backpropagation}\T\B & Rate & 99.42 & 1000 & 28 & 108.000 & 108.000 \\ \hline
    \end{tabular}
    \begin{tablenotes}
        \footnotesize
        \item[*] Scaled for 28nm process ($\times1.17$ for half a generation)
    \end{tablenotes}
    \end{threeparttable}
    }%
\end{table}

\begin{figure}[tb]
\centering
\subfloat[Accuracy]{%
  \includegraphics[clip,width=0.8\columnwidth]{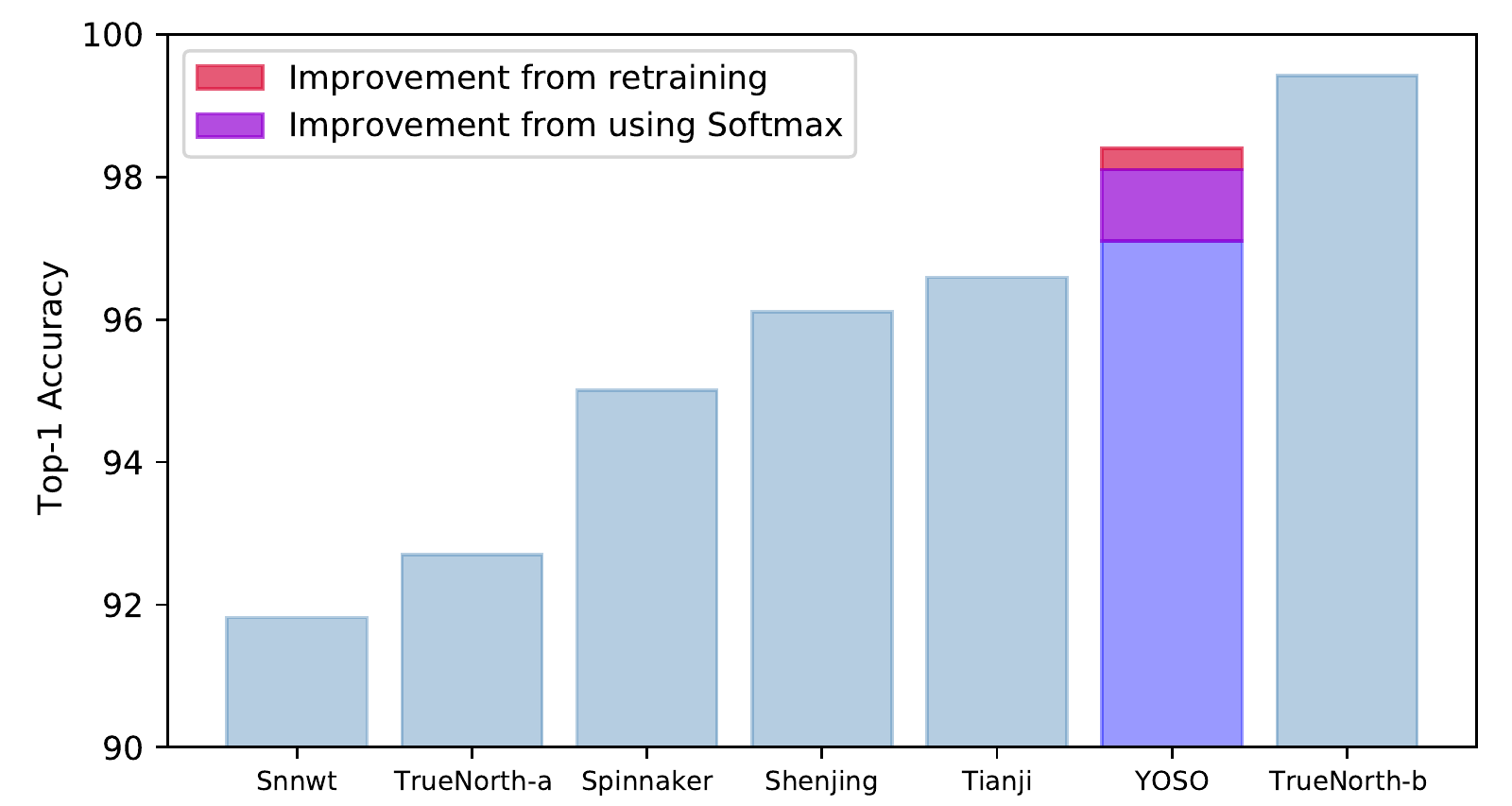}%
  \label{fig:accuracy_barchart}%
}

\subfloat[Power consumption]{
\includegraphics[clip,width=0.795\columnwidth]{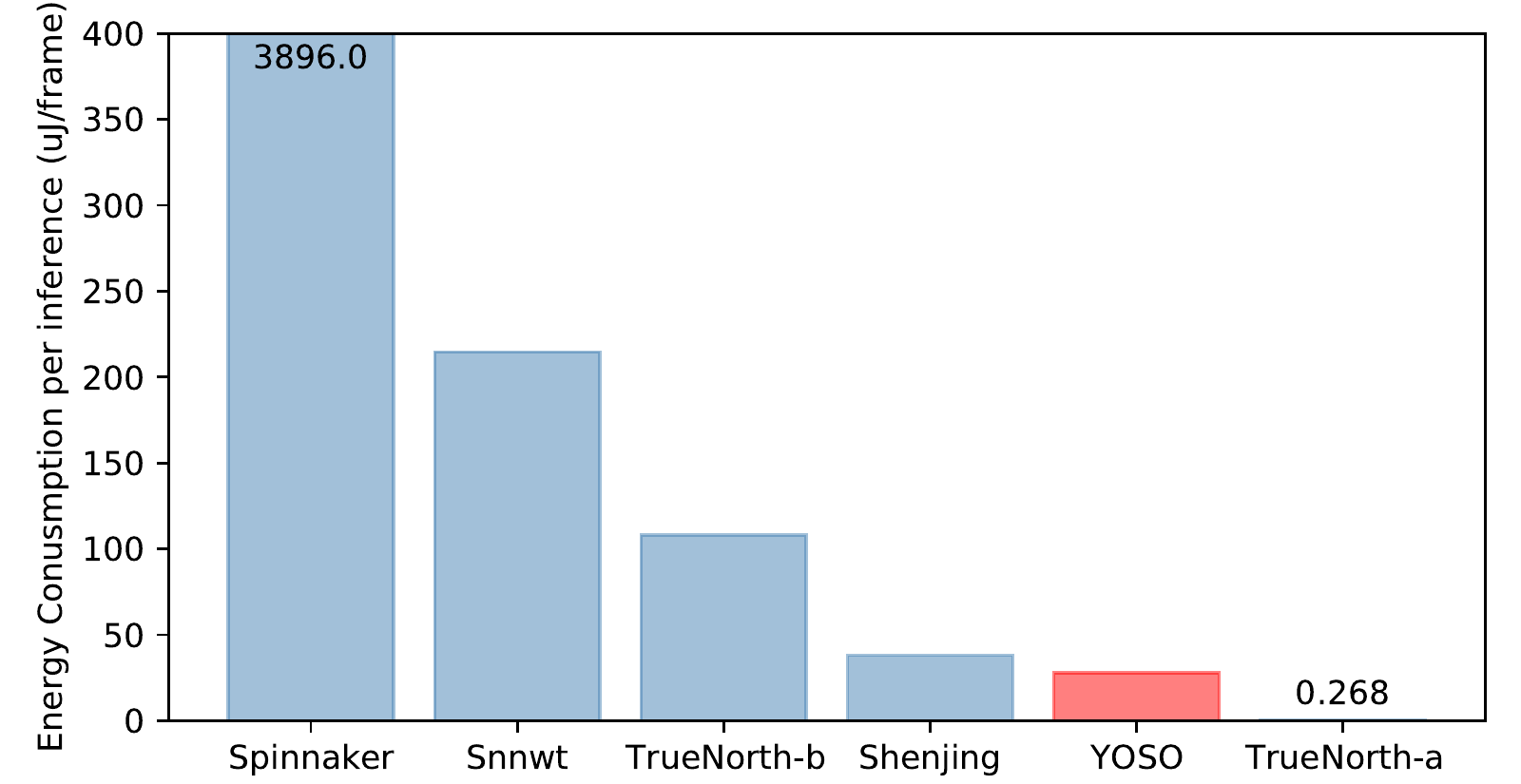}%
\label{fig:power_barchart}%
}
\caption{A comparison to other neuromorphic accelerators based on (a) accuracy and (b) power consumption. The YOSO accelerator is the only one that achieves both high accuracy and low power consumption at the same time. TrueNorth-a consumes almost $4\times$ the power to achieve its higher accuracy, while TrueNorth-b gives up a significant amount of accuracy (92.70\% vs. 98.40\% for our work) to achieve a lower power consumption.}
\end{figure}

Prior TTFS-encoding work~\cite{rueckauer_conversion_2018} has shown an accuracy of 98.30\%~(without quantization) on the MNIST dataset. Our work, in contrast, achieves 98.44\%~(without quantization) and 98.40\%~(with quantization) (See Table~\ref{table:mnist_results}, Figure~\ref{fig:accuracy_barchart}). Our proposed training method improves the accuracy of fully connected TTFS-encoded SNNs on the MNIST dataset compared to previous TTFS-encoded accelerators (by 0.85\%~\cite{DBLP:journals/corr/Mostafa16}) and most other rate-based SNN works. While the rate-based accelerator, TrueNorth-b, reaches a slightly higher accuracy than this proposed work, the costs are significant, with almost $4\times$ lower energy efficiency and a $129\times$ higher power consumption. The chosen input parameters to Algorithm~\ref{alg:finetuning-algorithm} were: (1) $|\{I^1...I^n\}| = 100$ (2) $\beta = 0.99$ (3) $\eta = 10$ (4) $\epsilon = 0.001$ (5) $K = 100$.

With the TTFS accelerator proposed in this work, power consumed per inference is significantly lower than that of the vast majority of rate-based accelerators, as shown in Table~\ref{tab:comparison_to_scalable_hardware} and Figure~\ref{fig:power_barchart}. Our work pushes the performance and efficiency boundary %
through the use of TTFS-encoded SNNs, and it is able to achieve both a high accuracy and low power simultaneously. Other works, like TrueNorth, sacrifice accuracy significantly to achieve a lower power consumption. While we demonstrate results on fully connected networks, our future work includes evaluating the performance gains obtained on larger networks and datasets~\cite{2009_cifar10,imagenet_2012}.

\subsection{Choice of output layer}

Typically, the output layer of an ANN is chosen to be a softmax layer because it ensures that the final layer's outputs are both normalized and strictly positive. For TTFS-encoded SNNs, the output neurons are one-hot encoded. One problem with such a method is if all neurons in the final layer receive negative inputs or inhibitory input comes later than another neuron spiking, either no neuron spikes or the incorrect neuron will spike.

One solution would be to perform softmax on the membrane potentials of the output layer neurons to determine the predicted output class, instead of choosing the neuron that spikes first. This method has allowed us to realize \textbf{1\%} improvement in accuracy of the network. This technique can be used when there is no need to use a purely spiking neural network.

\section{Conclusion}

In this work, we introduced the YOSO accelerator, and an improved Time-to-First-Spike training algorithm which demonstrates the viability of temporally-encoded SNNs for image classification tasks. %
To address the limitations of temporally-encoded SNNs, we proposed a novel training algorithm which achieves state of the art accuracy on temporally encoded SNNs. By combining this highly accurate temporal encoding method with our energy-efficient hardware design, YOSO, we demonstrate state-of-the-art temporal encoding results with high efficiency ($1.15\times$ better) and a low power consumption ($1.46\times$ better) over other state-of-the-art designs with comparable accuracy.

\section{Acknowledgements}

This research is supported by A*STAR under its RIE2020 IAF-ICP (Award I2001E0053).
Any opinions, findings and conclusions or recommendations expressed in this material are those of the authors and do not reflect the views of the A*STAR.

\renewcommand{\baselinestretch}{0.95}
\bibliographystyle{IEEEtran}
\bibliography{main}

\end{document}